\documentclass[twocolumn, switch]{article} % Method A for two-column formatting

\usepackage{preprint}

\usepackage{amsmath, amsthm, amssymb, amsfonts}

\usepackage[numbers,square]{natbib}
\usepackage[utf8]{inputenc}	% allow utf-8 input
\usepackage[T1]{fontenc}	% use 8-bit T1 fonts
\usepackage{xcolor}		% colors for hyperlinks
\usepackage[colorlinks = true,
            linkcolor = purple,
            urlcolor  = blue,
            citecolor = cyan,
            anchorcolor = black]{hyperref}	% Color links to references, figures, etc.
\usepackage{booktabs} 		% professional-quality tables
\usepackage{nicefrac}		% compact symbols for 1/2, etc.
\usepackage{microtype}		% microtypography
\usepackage{lineno}		% Line numbers
\usepackage{float}			% Allows for figures within multicol
\usepackage{tabularx}
\usepackage{algorithm,algpseudocode}
\usepackage{array}

\newcommand{\textgreek}[1]{\begingroup\fontencoding{LGR}\selectfont#1\endgroup}
\usepackage{textalpha} 

\usepackage[most]{tcolorbox} 
\tcbset{
  mygreybox/.style={
    breakable,              
    enhanced,               
    colback=gray!15,        % light-grey background
    colframe=gray!50,       % grey border
    boxrule=0.5pt,          % border thickness
    arc=3pt,                % rounded corners 
    left=1em, right=1em,    % inner horizontal padding
    top=1ex,  bottom=1ex,   % inner vertical padding
    before skip=\baselineskip,   % space before / after the box
    after skip=\baselineskip,
  }
}
\tcbset{
   promptstyle/.style={
    breakable,              % page-breaks automatically
    enhanced,               % required for breakable
    sharp corners,          % arc=3pt for rounded
    colframe=black!50,
    colback=gray!10,
    boxrule=.4pt,
    left=1em, right=1em,
    top=1ex, bottom=1ex,
    listing only,           % treat interior as verbatim
    listing options={       % control the verbatim listing
      basicstyle=\ttfamily\small,
      breaklines=true,
      breakautoindent=true,
      showstringspaces=false,
    },
  }
}
\usepackage{tikz}
\usepackage{amsmath}

\usetikzlibrary{
    shapes.geometric,
    arrows.meta,
    positioning,
    calc,
    fit
}

\usepackage{newfloat}
\DeclareFloatingEnvironment[name={Supplementary Figure}]{suppfigure}
\usepackage{sidecap}
\sidecaptionvpos{figure}{c}

\usepackage{titlesec}
\titlespacing\section{0pt}{12pt plus 3pt minus 3pt}{1pt plus 1pt minus 1pt}
\titlespacing\subsection{0pt}{10pt plus 3pt minus 3pt}{1pt plus 1pt minus 1pt}
\titlespacing\subsubsection{0pt}{8pt plus 3pt minus 3pt}{1pt plus 1pt minus 1pt}

\usepackage{xcolor,hyperref}

\definecolor{lime}{HTML}{A6CE39}
\DeclareRobustCommand{\orcidicon}{
	\begin{tikzpicture}
	\draw[lime, fill=lime] (0,0) 
	circle [radius=0.16] 
	node[white] {{\fontfamily{qag}\selectfont \tiny ID}};
	\draw[white, fill=white] (-0.0625,0.095) 
	circle [radius=0.007];
	\end{tikzpicture}
	\hspace{-2mm}
}
\foreach \x in {A, ..., Z}{\expandafter\xdef\csname orcid\x\endcsname{\noexpand\href{https://orcid.org/\csname orcidauthor\x\endcsname}
			{\noexpand\orcidicon}}
}

\title{DevNous: An LLM-Based Multi-Agent System for Grounding IT Project Management in Unstructured Conversation}

\usepackage{authblk}

\author[1\thanks{\tt{sdoropoulos@ihu.edu.gr}}]{Stavros Doropoulos\orcidA{}}
\author[1]{Stavros Vologiannidis\orcidC{}}
\author[2]{Ioannis Magnisalis\orcidB{}}

\affil[1]{Department of Computer, Informatics and Telecommunications Engineering, International Hellenic University, 62124 Serres, Greece; sdoropoulos@ihu.edu.gr; svol@ihu.gr}
\affil[2]{DG Informatics, European Commission, Brussels, Belgium; i.magnisalis@ihu.edu.gr}
\affil[ ]{\textit {Correspondence: sdoropoulos@ihu.edu.gr}}
\begin{document}

\twocolumn[ % Method A for two-column formatting
  \begin{@twocolumnfalse} % Method A for two-column formatting
  
\maketitle

\begin{abstract}
The manual translation of unstructured team dialogue into the structured artifacts required for Information Technology (IT) project governance is a critical bottleneck in modern information systems management. We introduce DevNous, a Large Language Model-based (LLM) multi-agent expert system, to automate this unstructured-to-structured translation process. DevNous integrates directly into team chat environments, identifying actionable intents from informal dialogue and managing stateful, multi-turn workflows for core administrative tasks like automated task formalization and progress summary synthesis. To quantitatively evaluate the system, we introduce a new benchmark of 160 realistic, interactive conversational turns. The dataset was manually annotated with a multi-label ground truth and is publicly available. On this benchmark, DevNous achieves an exact match turn accuracy of 81.3\% and a multiset F1-Score of 0.845, providing strong evidence for its viability. The primary contributions of this work are twofold: (1) a validated architectural pattern for developing ambient administrative agents, and (2) the introduction of the first robust empirical baseline and public benchmark dataset for this challenging problem domain.
\end{abstract}
%\keywords{First keyword \and Second keyword \and More} % (optional)
\vspace{0.35cm}

  \end{@twocolumnfalse} % Method A for two-column formatting
] % Method A for two-column formatting

%\begin{multicols}{2} % Method B for two-column formatting (doesn't play well with line numbers), comment out if using method A

%%%%%%%%%%%%%%%  Main text   %%%%%%%%%%%%%%%
% \linenumbers

\section{Introduction}
\begin{figure*}[htbp]
    \begin{center}

    \includegraphics[width=0.99\textwidth]{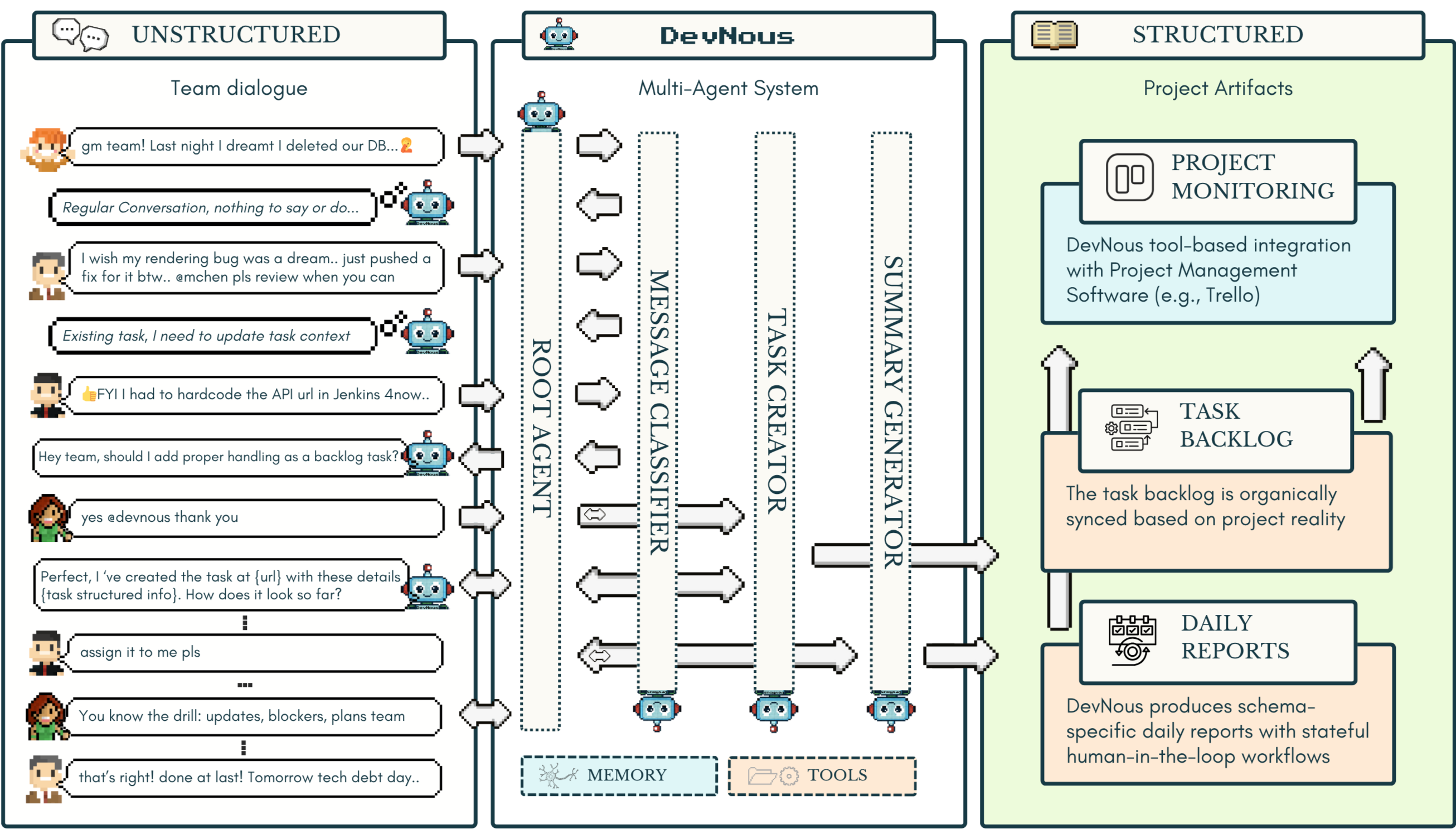}
\end{center}
\scriptsize{The conceptual architecture of the DevNous system, illustrating its role as a bridge between unstructured team dialogue and structured project artifacts. The \texttt{root agent} orchestrates specialized agents to ingest, classify, and statefully act upon conversational data, ultimately maintaining formal artifacts like the task backlog and progress updates. }
    
    \caption{DevNous Multi-Agent System}
    \label{fig:devnous-architecture}
\end{figure*}
The effective management of Information Technology (IT) projects is a critical determinant of organizational success, yet such projects are notoriously susceptible to failure~\cite{karnouskos2024relevance, schwalbe2016information}. This persistent challenge has spurred a reciprocal dynamic where project management needs drive technological innovation, and new technologies are deployed to enhance project execution and communication~\cite{anantatmula2008role}. Extensive research has documented the application of Machine Learning (ML) and Deep Learning (DL) models to project management challenges, establishing their strategic value~\cite{gil2021application,taboada2023artificial,temitope2020software}.  However, these tools typically function as ad hoc analytical systems, lacking deep integration into the primary collaboration and communication workflows where much of a project's state is implicitly defined and negotiated. 

This challenge is amplified by the operational reality of modern IT projects. Formal project management frameworks mandate the maintenance of artifacts such as task backlogs and progress reports, which are essential for tracking and control across knowledge areas like scope, schedule, and communication~\cite{pmi2021pmbok}. However, in contemporary practice, particularly with the rise of distributed teams reliant on chat-based platforms~\cite{dingel2020many}, the information required to maintain these artifacts originates from high-volume, unstructured conversations. The project manager, whose role has evolved into that of a servant-leader and facilitator~\cite{gasemagha2021project, greenleaf2013servant}, bears the cognitive load of monitoring these dialogues, identifying emergent tasks, and manually translating informal requests into structured data and insights. This manual process is not only resource-intensive but is also inherently prone to error and omission, directly contributing to the managerial shortcomings that precipitate project failure~\cite{karnouskos2024relevance}. A critical gap therefore exists between the informal, dynamic nature of team collaboration and the structured data required for effective project governance.

The emergence of Generative AI offers a path beyond this limitation, enabling autonomous agents capable of operating directly within these core communicative processes. The recent advent of Large Language Models (LLMs) has catalysed a paradigm shift, enabling machines to comprehend, generate, and reason with human language at an unprecedented level of sophistication. Despite this potential, the integration of LLMs as autonomous agents within IT project workflows remains underexplored, presenting a compelling opportunity for new research. The primary challenge lies in translating the unstructured, informal, noisy, and often fragmented discourse of team collaboration into structured, actionable project data and operations. Motivated by this challenge, this paper proposes \textbf{DevNous} (from \textit{dev}elopment and the greek word \textgreek{νοῦς}, meaning mind or intellect), a hierarchical multi-agent system designed to operate autonomously within communication-centric project environments. Our central hypothesis is that an intelligent agent system can successfully augment project execution by actively monitoring communication channels, interpreting conversational intent, and executing administrative and logistical tasks. DevNous's architecture is designed to first classify incoming human messages for relevance and then trigger specific agentic workflows. To validate this framework, we developed and evaluated a prototype system focusing on two core, high-frequency and artifact-heavy activities in IT projects. The primary contributions of this work are twofold:
\begin{enumerate}
    \item \textbf{A Multi-Agent Architecture for LLM-based Project Management:} We present and validate a novel architecture for autonomous agents in dialogue-based project management. The proposed system, DevNous, uses functional specialization to classify actionable intents from informal team dialogue and orchestrate stateful, multi-turn workflows for core administrative tasks, including automated task formalization and progress report synthesis.  This design avoids the fragility of a monolithic agent while limiting the potential coordination and computational overhead associated with larger-scale agent swarms~\cite{kim2025cost}.

    \item \textbf{A Rigorous Evaluation Framework and Empirical Baseline:} We introduce a comprehensive methodology for evaluating this class of interactive agents. This includes a new, 160-turn benchmark dataset of path-dependent dialogues generated by an interactive Synthetic Generation Agent (SGA) and annotated by human annotators using a multi-label annotation schema. The dataset is publicly available at \url{https://doi.org/10.5281/zenodo.16755864}. Our experiments, conducted using this framework, show that DevNous achieves a strong exact match turn accuracy of 81.3\%, providing the first robust empirical baseline for this problem domain.
\end{enumerate}
The proposed multi-agent architecture is validated through a comparative evaluation where we demonstrate that DevNous significantly outperforms a monolithic agent baseline.
The remainder of this paper is organized as follows: Section 2 provides background, and Section 3 reviews related work. Sections 4 and 5 describe our proposed methodology and evaluation framework. Section 6 presents our results while Section 7 discusses our findings. Sections 8 and 9 present the study’s limitations and conclude the paper, outlining directions for future research.

\section{Background}
Our research is at the intersection of project management methodologies, the application of AI to project management, and the design of LLM-agent architectures. This section provides background on these foundational domains to position and contextualize our proposed contribution.

\subsection{Project Management Methodologies}

Project management has historically established a paradigm of utilizing structured artifacts as the primary medium for project execution and control. This formalization began from foundational contributions for structured representations of work with early visual scheduling tools like the Gantt chart~\cite{gantt1919organizing} and core management theories defining functions such as planning and coordinating~\cite{fayol1949general}. Sophisticated analytical techniques like the deterministic Critical Path Method (CPM)~\cite{kelley1961critical} and the probabilistic Program Evaluation and Review Technique (PERT)~\cite{malcolm1959pert} were proposed for managing schedule complexity. Additionally, project management organizations like the AACE~\cite{aace_history} and the Project Management Institute (PMI)~\cite{pmi_history} contributed to this formalization. The introduction of artifact-heavy best practices within standards like the PMBOK~\cite{pmi2021pmbok} and PRINCE2~\cite{office2009managing} further reinforced this principle. This paradigm is evident in traditional approaches where project progression is defined by sequential artifacts~\cite{royce1970managing}, but is also apparent, in non-static structure, in Agile~\cite{sirisha2024project} and Hybrid methodologies~\cite{reiff2022hybrid}. The core legacy of this evolution is the systemic expectation that project status and success are driven and defined by structured~documentation.

The inherent rigidity of traditional approaches proved challenging in dynamic environments, prompting a shift toward more adaptive methodologies. The artifact-centric paradigm encountered significant mutations when faced with the ever-changing requirements of modern IT and software development. The  methodological shift that was catalysed by the Agile Manifesto~\cite{beck2001manifesto} prioritized iterative value creation, flexibility, and continuous team interaction and feedback over static documentation and processes. Frameworks such as Scrum~\cite{schwaber1997scrum}, Kanban~\cite{anderson2010kanban}, and  Extreme Programming (XP)~\cite{beck2000extreme}, alongside related methodologies like Lean~\cite{womack2007machine} and Scaled Agile Framework (SAFe)~\cite{leffingwell2007scaling}, operationalized this philosophy by augmenting and building the project artifacts with insights from the high-frequency conversations, interactions, and collaborative activities of the stakeholders. At the same time, teams' way of work was affected by the widespread adoption of chat-based platforms for remote collaboration~\cite{dingel2020many}. The convergence of these trends has created a profound operational dissonance: a persistent need for structured artifacts for governance must now be satisfied by manually interpreting a high-volume, unstructured, and continuous stream of team dialogue.  These transformations emerged together with the redefinition of the project manager's role from a process controller to that of a facilitator of this process~\cite{greenleaf2013servant, gasemagha2021project, meng2017role, coetzer2017functions}. This manual ``unstructured-to-structured'' translation represents the critical bottleneck of administrative overhead and data loss that this research aims to address through agentic automation.

\subsection{Artificial Intelligence in Project Management}

The application of Artificial Intelligence (AI) to augment project management processes is a well-researched domain, with numerous systematic reviews detailing its broad impact~\cite{gil2021application, taboada2023artificial, hashfi2023exploring}. 
Scholarly research has proposed a diverse collection of models designed to optimize several project phases including initiation, planning, control and execution. Significant research effort has been dedicated to predictive analytics, particularly in areas like project success, cost~\cite{ mir2021neural} and effort estimation~\cite{nassif2013towards}. Other prominent application domains include duration forecasting~\cite{irfan2011planning}, risk management~\cite{yet2016bayesian}, resource optimization ~\cite{koulinas2014particle}, project performance monitoring~\cite{Kim2015} and adaptive digital twins~\cite{Hribernik2021}. The adoption of AI technologies in project management is enhancing decision making and project execution success. However, the unifying characteristic of these applications is that they are designed as specialized analytical models that operate on well-defined problem scopes. While some may parse natural language text, they fundamentally lack the generalizable, cross-functional reasoning required to interpret the ambiguous and dynamic nature of chat-based discourse. Consequently, these systems are unable to address the main challenge of orchestrating work as it emerges from dynamic team dialogue. This capability gap necessitates a new class of AI grounded in general-purpose language understanding.

\subsection{The Emergence of LLM-Based Agents}

The new class of AI capable of general-purpose language understanding emerged from a paradigm shift within Natural Language Processing (NLP). This evolution progressed from early statistical language models~\cite{jelinek1998statistical, rosenfeld2000two}, through the development of distributed word representations that mitigated the curse of dimensionality~\cite{bengio2003neural, mikolov2013efficient}, and recurrent memory-enabled architectures like Long Short-Term Memory (LSTMs)~\cite{hochreiter1997long}. The critical breakthrough, however, was the development of the Transformer architecture, which replaced recurrence with a self-attention mechanism, enabling unprecedented scalability and performance on long-range dependencies~\cite{vaswani2017attention}. This architecture paved the way for large-scale pre-training, producing models like BERT and GPT that learn general language representations from massive datasets~\cite{devlin2019bert, brown2020language}. The application of scaling laws~\cite{kaplan2020scaling} and alignment techniques like Reinforcement Learning from Human Feedback (RLHF)~\cite{ouyang2022training} has resulted in today's LLMs  that exhibit general-purpose reasoning and emergent problem-solving abilities far exceeding their predecessors~\cite{wei2022emergent, raiaan2024review}.

The generalizable reasoning capabilities of these LLMs have enabled the development of a new generation of agents, realizing a long-standing vision in AI of systems possessing properties of autonomy, reactivity, and social ability~\cite{wooldridge1995intelligent, wang2024survey}. Unlike specialized models, these LLM-based agents can perform complex, multi-step tasks by dynamically drafting and executing plans using input from their environment. These agents typically involve modules for defining a profile (role, persona), handling memory (context), planning (task decomposition and orchestration), and taking actions (tool use)~\cite{wang2024survey}. 

The behavior of an agent is shaped by its profile, often specified via a handcrafted prompt. Research has shown that assigning specific roles and personas  significantly improves performance in multi-agent systems by providing traits, focus, and constrains~\cite{park2023generative, chen2023agentverse, qian2023chatdev}. The agent's behavior is also directly influenced by all the available information in the provided context. To overcome the limitations of limited context windows, agents can utilize memory mechanisms. These are often hybrid, combining short-term memory like the recent conversation history with long-term memory stores for information archival and retrieval, which is what we use in our proposed architecture.

Planning is crucial for agent autonomy and task execution accuracy. Early methods relied on simple in-context prompting, which was significantly improved by Chain-of-Thought (CoT) prompting that elicits step-by-step reasoning from the LLM~\cite{wei2022chain}. The current state-of-the-art has moved towards more dynamic and interactive planning like the ReAct framework that demonstrated the importance of combining reasoning with action and observation, allowing agents to adapt their plans based on environmental feedback~\cite{yao2023react}. Our work adopts a ReAct-style feedback loop but constrains it within a structured, hierarchical coordination suitable for our domain.

Finally, agents derive much of their capabilities from the use of external tools. This can range from calling APIs and executing code to querying databases or using Retrieval-Augmented Generation (RAG) mechanisms~\cite{gao2024retrievalaugmentedgenerationlargelanguage}. Seminal work in this area includes frameworks that teach LLMs to use tools by observing examples~\cite{qin2023toollm} and systems that orchestrate multiple specialist models as tools~\cite{shen2023hugginggpt}. The ability to ground actions in a reliable set of tools is critical for mitigating hallucination and allowing agents to affect external environments~\cite{10.1145/3704435}. The use of tools is particularly important for our ``unstructured-to-structured'' translation for achieving schema-based translations and syncing the project's reality with and from external project management software.

\section{Related Work}

In the domain of project management, scholarly work has only recently begun to explore LLMs mainly though a non-agentic perspective. Several studies have focused on the human factors of LLM adoption, examining the  dynamics and user acceptance of AI assistants in Agile meetings~\cite{cabrero2024exploring} or the organizational factors influencing a project manager's appropriation of generative AI tools~\cite{felicetti2024artificial}. Other research has investigated the application of LLMs to well-bounded PM tasks, such as automating the generation of user stories or creating reports for Agile ceremonies~\cite{bahi2024integrating, alliata2024ai}. While valuable, these studies predominantly treat LLMs as powerful assistants for specific tasks, rather than as autonomous agents deeply integrated into the team's operational workflow.

A more involved agent-based approach is evident in the domain of software engineering, where the distinction between standalone LLMs and tool-using agents is well defined~\cite{jin2024llms}. The potential of agent-based design approach is evidenced by the success of multi-agent systems in systems such as ChatDev~\cite{qian2023chatdev} and frameworks like AutoGen~\cite{wu2023autogen} that can autonomously perform complex development tasks. These multi-agent systems have been designed to autonomously handle development lifecycles, with agents assuming roles like managers, programmers, and testers to coordinate on tasks such as code generation from high-level requirements~\cite{zhang2024experimenting, qian2023chatdev}.

In the PM domain, agent-based recent work utilizes LLM-Swarms to examine their effect in agile project management by coordinating agent clusters representing key roles such as managers and engineers~\cite{hussain2024collective}. Another approach is seen in simulation frameworks like CogniSim, which evaluates agent interactions based on established enterprise practices such as SAFe~\cite{cinkusz2024cognitive}.
However, these agent-based systems are designed to operate on well-structured problem definitions, such as formal requirements, scenarios or code repositories. They are not architected to handle high-velocity conversational environments of an IT project team. Consequently, a research gap persists at the intersection of these fields. To our knowledge, no prior work has presented the design, implementation and evaluation of a multi-agent system specifically architected to autonomously perform coordinative and administrative tasks by interpreting the informal dialogue of live collaborative environments.

\section{Methodology}

To address the research gap identified in the literature, we designed and implemented \textbf{DevNous}, a multi-agent expert system for autonomous project management within unstructured conversation environments. This section details the formal problem definition, the system's architecture, and the methodology underpinning its operation, which was validated through the implementation of two critical and high-overhead agile project workflows: automated task formalization and automated progress synthesis.

\subsection{Problem Formulation}
We formally define the problem of autonomous project management in a chat-based, unstructured conversation environment as a sequential decision-making process. 
The system operates on a continuous stream of messages $m_1, m_2, \dots, m_t$, where each message $m_i$ is a tuple $(c_i, u_i, \tau_i)$ representing the content $c_i$, user $u_i$, and timestamp $\tau_i$. At any time $t$, the system's knowledge is captured in a state $s_t = (B_t, H_t, W_t)$, which includes the project artifact repository $B_t$ (e.g., the task backlog and team members information), the recent conversation history $H_t$, and the active workflow state $W_t$. The core challenge is to learn a policy, $\pi(\cdot)$, that maps the current state $s_t$ and a new incoming message $m_{t+1}$ to an optimal action $a_{t+1}$ from a predefined action space $\mathcal{A}$:
\begin{equation}
    a_{t+1} = \pi(s_t, m_{t+1})
\end{equation}
The action space $\mathcal{A}$ includes high-level coordinative and administrative functions, such as:
\[
\mathcal{A} = \left\{
\begin{aligned}
&\texttt{CREATE\_TASK},\\
&\texttt{CONTINUE\_WORKFLOW},\\
&\texttt{UPDATE\_CONTEXT},\\
&\texttt{GENERATE\_SUMMARY},\\
&\texttt{NO\_ACTION}
\end{aligned}
\right\}.
\]
The primary difficulty lies in the nature of $m_{t+1}$, where the content $c_{t+1}$ is unstructured natural language. Therefore, the policy $\pi(\cdot)$ must perform complex semantic interpretation. DevNous is a hierarchical multi-agent system designed to implement this policy, with LLMs serving as the core reasoning engine for interpreting $(s_t, m_{t+1})$ and selecting an appropriate action.

\begin{table*}[htbp]
\footnotesize
\caption{DevNous Agent Architecture: Roles, Personas, and Mechanisms.\label{tab:agent_architecture}}
\begin{tabularx}{\textwidth}{p{2cm} >{\raggedright\arraybackslash}p{2.5cm} XX}
\toprule
\textbf{Agent} & \textbf{Role / Persona} & \textbf{Core Responsibility} & \textbf{Primary Mechanism / Output} \\
\midrule
\texttt{Root Agent} & Orchestrator / \mbox{Supervisor} & Message triage and sub-agent delegation based on context, message category, and intent. & Delegation: Executes the policy ($\pi(\cdot)$) that maps classified intent to a specific sub-agent transfer. Executes Sub-agent transfers. \\
\addlinespace
\texttt{Classifier} & Intent Analyzer / \mbox{Systematic Analyst} & Classify user messages into a predefined set of intents (Category and Action tuples). & Intent Classification. Returns structured JSON output corresponding to an action in $\mathcal{A}$ with confidence score. \\
\addlinespace 
\texttt{Task Creator} & Task Facilitator / \mbox{Interactive}  & Manage a human-in-the-loop (HITL) process to formalize unstructured requests into tasks. & Stateful HITL: unstructured-to-structured translator with persistent workflow states ($W_t$). Outputs structured tasks. \\
\addlinespace
\texttt{Summary Generator} & Report Synthesizer / \mbox{Factual Observer} & Analyze conversation history and project data to generate daily progress reports. & Conversational Data Synthesis: Retrieves, analyzes, and aggregates data from $H_t$ and $B_t$, producing a structured textual summary. Generates structured reports for team members. \\
\bottomrule
\end{tabularx}
\end{table*}
\subsection{System Architecture}
\label{methodology-section}

The system's policy $\pi(\cdot)$ is implemented with a hierarchical multi-agent architecture that decomposes the main policy $\pi(\cdot)$ into three distinct sub-policies, each implemented by a specialized agent, as depicted in Figure \ref{fig:devnous-architecture}. A central root agent functions as orchestrator, receiving all incoming messages and delegating tasks to a team of specialized sub-agents based on the output of the message classification agent. The root agent orchestrates three specialized sub-agents: a) the message classifier sub-agent for actionable intent recognition, b) the task creator sub-agent for backlog management, and c) the summary generator sub-agent for report synthesis.

The classifier agent implements the actionable intent recognition policy ($\pi_{\text{classify}}$) which maps the input state and message to the highest-confidence (Category,Action) tuple it identifies. Based on this output the primary action to be taken at each turn is decided:
\begin{equation}
    (k_{t+1}, a_{t+1}) = \pi_{\text{classify}}(s_t, m_{t+1}).
\end{equation}
Here $k_{t+1}\in\mathcal{K}$ is the predicted category and $a_{t+1}\in\mathcal{A}$ is the action, where
\[
\mathcal{K} = \left\{
\begin{aligned}
&\texttt{NEW\_TASK},\\
&\texttt{EXISTING\_TASK},\\
&\texttt{WORKFLOW\_RESPONSE},\\
&\texttt{SUMMARY\_TRIGGER},\\
&\texttt{REGULAR\_CONVERSATION}
\end{aligned}
\right\}.
\]

The root agent based on $\pi_{\text{classify}}$ conditionally invokes the task creator policy ($\pi_{\text{task}}$) or summary generator policy ($\pi_{\text{summary}}$). 
These sub-agents implement stateful workflow policies that may span multiple interactive turns.

The task formalization policy ($\pi_{\text{task}}$), executed by the task creator agent, accepts the message and the current state $(s_t, m_{t+1})$, and produces, through a human-in-the-loop dialogue,  an updated artifact repository $B'$, where a new task has been added:
\begin{equation}
    B' = \pi_{\text{task}}(s_t, m_{t+1}).
\end{equation}
The summary generation policy ($\pi_{\text{summary}}$), executed by the summary generator agent, takes as input the current state and message $(s_t, m_{t+1})$ and produces, through a human-in-the-loop process, a structured, textual summary:
\begin{equation}
    \text{Summary} = \pi_{\text{summary}}(s_t, m_{t+1}).
\end{equation}

The full orchestration policy $\pi(\cdot)$ is therefore a conditional hierarchical composition of these sub-policies.

The core operational logic of this hierarchy is detailed in the message handling workflow, presented in Algorithm \ref{alg:policy_implementation}. This control loop illustrates how the system triages incoming conversational messages and routes control to the appropriate specialized agent based on classified actionable intent and current workflow state. The return statement transfers control and yields either (a) no outward message (\textit{null}), (b) the user response and the updated state from the invoked sub-policy (e.g., the updated backlog $B'$ from $\pi_{\text{task}}$ or the $\text{Summary}$ from $\pi_{\text{summary}}$), or (c) a custom generated response (\texttt{GenerateResponse}) from the root policy. To ensure proper unstructured-to-structured translation all sub-agent outputs are validated against predefined JSON Schemas which are defined in Tables~\ref{tab:schema-messageclassification},~\ref{tab:schema-task},~\ref{tab:schema-summary} of Appendix~\ref{appendix:schemas}.

\begin{algorithm}[htbp]
\caption{DevNous Orchestration Policy $\pi(\cdot)$}
\label{alg:policy_implementation}
\begin{algorithmic}[1]
\footnotesize
\Procedure{ExecutePolicy}{$s_t, m_{t+1}$}
    \State \tiny\Comment{$s_t$: current state $(B_t, H_t, W_t)$; $m_{t+1}$: new message}\footnotesize
    \State $(k_{t+1}, a_{t+1}) \gets \pi_{\text{classify}}(s_t, m_{t+1})$  \tiny\Comment{Implemented by the \texttt{Classifier Agent}}\footnotesize
    \If{$a_{t+1} \in \{\texttt{NO\_ACTION}, \texttt{UPDATE\_CONTEXT}\}$}
        \State \Return \textit{null} \tiny\Comment{Passively update state without responding}\footnotesize
    \ElsIf{$a_{t+1} = \texttt{CREATE\_TASK}$}
        \State \Return $\pi_{\text{task}}(s_t, m_{t+1})$
    \ElsIf{$a_{t+1} = \texttt{GENERATE\_SUMMARY}$}
        \State \Return $\pi_{\text{summary}}(s_t, m_{t+1})$
    \ElsIf{$a_{t+1} = \texttt{CONTINUE\_WORKFLOW} \land s_t.W_t\texttt{.isActive}$}
        \If{$s_t.W_t\text{.type} = \texttt{task\_workflow}$}
            \State \Return $\pi_{\text{task}}(s_t, m_{t+1})$
        \ElsIf{$s_t.W_t\text{.type} = \texttt{summary\_workflow}$}
            \State \Return $\pi_{\text{summary}}(s_t, m_{t+1})$
        \EndIf
    \Else
        \State \Return \Call{GenerateResponse}{$s_t, m_{t+1}$}
    \EndIf
\EndProcedure
\end{algorithmic}
\end{algorithm}

\subsection{Multi-Agent Team Configuration}
The per agent functional separation in our architecture is achieved through role-defining profiles and instructions, functionality based tool actions, memory access and cross agent planning. Each agent’s behavior is governed by a handcrafted profile, implemented as a detailed instruction prompt. These prompts go beyond simple persona assignment to include behavioral constraints, workflow logic, and guidelines for tool use as detailed in Table~\ref{tab:agent_architecture}, a technique shown to improve multi-agent performance~\cite{chen2023agentverse}. The complete set of instructions is presented in Appendix~\ref{appendix:agents-prompts}.

Agent actions within the DevNous framework are executed through a tool-based abstraction layer that the agents interact with and use to update the state ($s_t$). The LLM produces structured calls to a set of registered Python functions, categorized by functionality into four primary tool classes: (1) \textit{Memory tools}, responsible for managing session state (e.g., \texttt{memorize\_string}, \texttt{get\_conversation\_history}); (2) \textit{Communication tools}, which interface with chat systems for message ingestion and dispatch (e.g., \texttt{process\_message}, \texttt{send\_message}); (3) \textit{Project management tools}, which bridge the LLM with external task-tracking platforms (e.g., \texttt{get\_tasks}, \texttt{update\_task}); and (4) \textit{Workflow tools}, which support human-in-the-loop operations by managing process state transitions (e.g., \texttt{start\_workflow}, \texttt{end\_workflow}). The full list of tools is detailed in Table~\ref{tab:devnous-tools} of the Appendix~\ref{appendix:tools}. The agents have access to these tools based on the principle of least privilege~\cite{leastpriviledge}. The separation of access to tools and processes acts as a security mechanism, with agents having authorization over a specified subset of the action and memory space. 

DevNous implements a hybrid memory architecture to maintain proper context state across multi-turn interactions. Short-term memory consists of the recent conversation history  ($H_t$) while its long-term memory utilizes a persistent state managed through dedicated tools, with information such as the active workflow status ($W_t$), team member data, and a cache of tasks ($B_t$) retrieved from external systems.

The system's planning methodology employs a hybrid strategy that combines structured delegation-based coordination pattern with dynamic agent transfers and tool execution. A key heuristic for stability is that sub-agents are disallowed from peer-to-peer transfers and must return control to the root agent after task execution. At the sub-agent level, execution is adaptive, based on a feedback-driven methodology inspired by the ReAct framework's thought-action-observation triplets~\cite{yao2023react}, allowing agents to dynamically adjust their behavior based on tool outputs and user responses. 

The specific orchestration of these components to execute the two primary workflows of automated task formalization and progress synthesis is detailed in Appendix~\ref{appendix:subages-architecture}.

\subsection{Handling Conversational Ambiguity}
A primary methodological contribution of DevNous is its ability to operate robustly within noisy, unstructured chat environments. This is achieved through two key heuristics engineered into the agent profiles and the central orchestration logic and thus affecting the system's policy ($\pi(\cdot)$):
\begin{enumerate}
    \item \textbf{Cross-Talk Detection:} Agents are explicitly instructed to differentiate between messages directed at them and general team conversation (cross-talk). The classifier agent enables the root agent to passively observe instead of intrusively responding to incoming messages.
    \item \textbf{Workflow State Continuity:} To prevent contextual drift during multi-turn interactions, the active workflow state ($W_t$) is a primary input for reasoning. The classifier is instructed to interpret new messages as possible responses to an active workflow to maintain conversational coherence.
\end{enumerate}

\section{Evaluation Framework}

\begin{table*}[htbp]
\centering
\caption{Definitions for message category and action annotation labels}
\label{tab:category_definitions}
\footnotesize
\begin{tabularx}{\textwidth}{@{}l >{\raggedright\arraybackslash}X@{}}
\toprule
\textbf{Message Category} & \textbf{Definition} \\
\midrule
\texttt{NEW\_TASK} & Assigned to the first message in a thread that proposes a new, untracked task. Untracked tasks are those not present in the Backlog ($B_t$) or in the recent conversation ($H_t$).\\
\addlinespace
\texttt{EXISTING\_TASK} & A message referencing a tracked task. \\
\addlinespace
\texttt{WORKFLOW\_RESPONSE} & Assigned to messages that are direct responses to an agent's or user's question within an active, multi-turn workflow (e.g., task creation or summary confirmation). \\
\addlinespace
\texttt{REGULAR\_CONVERSATION} & Assigned to cross-talk, acknowledgements, or discussions not containing actionable project management information. \\
\addlinespace
\texttt{SUMMARY\_TRIGGER} & Assigned to commands or contextually infered need to generate a summary report. \\

\toprule
\textbf{Action Type} & \textbf{Definition and Typical Pairing} \\
\midrule
\texttt{CREATE\_TASK} & Initiate a workflow to create a new task. Typically paired with NEW\_TASK Message Category. \\
\addlinespace
\texttt{UPDATE\_CONTEXT} & Silently observe and update information for a previously established topic. Typically paired with EXISTING\_TASK Message Category. \\
\addlinespace
\texttt{CONTINUE\_WORKFLOW} & Take the next step in an active, multi-turn workflow. Typically paired with WORKFLOW\_RESPONSE Message Category. \\
\addlinespace
\texttt{NO\_ACTION} & Do not respond or record information. Typically paired with REGULAR\_CONVERSATION Message Category. \\
\addlinespace
\texttt{GENERATE\_SUMMARY} & Initiate the summary generation workflow. Typically paired with SUMMARY\_TRIGGER Message Category. \\
\bottomrule
\end{tabularx}
\end{table*}
The evaluation of LLM-based agents is an open research problem, as their interactive and dynamic nature presents challenges that cannot be fully captured by traditional benchmarks~\cite{wang2024survey}. A critical shortcoming of existing approaches is a focus on task-level accuracy that often neglects the path-planning complexities of real-world conversational workflows, leading to a  disconnection between theoretical performance and practical effectiveness~\cite{kapoor2024ai}. This shortcoming dictates the necessity of developing an evaluation methodology that truly reflects the performance of proposed agentic architectures. While various objective and subjective evaluation methods exist, the foundation of any rigorous quantitative analysis is a high-quality, domain-specific benchmark dataset.

However, in the domain of IT project management, there is a distinct lack of publicly available datasets that capture the informal, multi-threaded nature of chat-based team collaboration. Existing datasets from software development communities, such as the DISCO dataset~\cite{discodataset}, were examined but found to be unsuitable. These datasets, while valuable for a plethora of analysis and research problems, do not contain the specific coordinative and administrative speech flows (e.g., task creation from informal requests, contextual updates, managerial discussions, summary generation requirements) that are the primary focus of our work. Additionally, to fully evaluate our system the interaction of an IT team with the agent would be required and thus static datasets are insufficient. This gap necessitated the creation and annotation of a new benchmark dataset to enable a valid evaluation of the DevNous system. The following subsections detail the dataset generation, annotation schema, the evaluation protocol, the specific metrics used in this evaluation, as well as implementation details.

\subsection{Dataset Generation}
Procuring informal, project-specific internal chat exchanges from active IT teams, along with data on their members and task backlogs, is challenging because of privacy and intellectual-property concerns. To address the lack of this required interactive dataset, we adopted a synthetic data generation approach. This methodology has been found effective for creating datasets for intent classification tasks, especially when using larger LLMs~\cite{maheshwari2024efficacy}. Rather than creating data using a static single LLM query, we engineered SGA, a separate, context-aware conversational agent tasked with performing a realistic turn-based interaction of a simulated team with DevNous. 

The SGA was powered by Anthropic's Claude-Sonnet 3.7~\cite{claude_sonnet}, a state-of-the-art language model that we have found to perform well in context-aware language understanding~\cite{doropoulos2025beyond}. The instruction set we used provides the SGA with a team persona and the team members info together with the initial list of tasks backlog ($B_t$). The complete prompt used for the data generation is presented in Appendix~\ref{appendix:agent-sga-prompts}. The agent was instructed to follow realistic conversational patterns and constraints derived from real-world developer chats. The core of the methodology is a dynamic, turn-based interaction loop:
\begin{enumerate}
    \item The SGA generates a human-simulated message based on the recent conversational history, including DevNous's previous responses.
    \item This message is sent to the DevNous system, which processes it and generates a response.
    \item The full interaction (human input and DevNous output) is then added to the conversational context.
    \item The updated context is used by the SGA to generate the next, human-simulated message.
\end{enumerate}
This interactive process was repeated to generate a benchmark dataset consisting of 8 independent, multi-turn conversational dialogues. Each dialogue contains 20 human-simulated turns, for a full benchmark of 160 human turns and 169 message classifications (category and action tuples). This approach ensures the resulting dialogues are not only realistic but also path-dependent, providing a challenging, realistically simulated evaluation dataset for our system. The complete, annotated benchmark dataset is openly available in Zenodo at \begin{center}
    \url{https://doi.org/10.5281/zenodo.16755864}.
\end{center}
\subsection{Annotation Process}

We followed a turn-based, multi-label annotation methodology. Each of the 160 human-simulated turns was manually annotated with a multiset of actionable intent (MessageCategory, Action) tuples, allowing for duplicates if required. Each tuple represents the message's category, function or goal and the action ($a_{t+1}$), which defines the ideal step the agent should plan to take in response to the message ($m_{t+1}$) and the current state ($s_t$) based on their instructed policy ($\pi(\cdot)$). The definitions for the five possible message category and the five action labels,  along with their typical pairings, are detailed in Table~\ref{tab:category_definitions}.

To ensure the reliability and objectivity of this annotation process, we conducted an inter-annotator agreement study. A second, independent annotator was trained on the definitions outlined in our schema and tasked with labeling a randomly selected subset of 40 turns (25\% of the dataset). We then measured the agreement between the two sets of annotations using the multiset F1-score (see Eq.\ref{eq:multiset_f1_agreement}), which is well-suited for this multi-label, multiset tasks.  The analysis yielded a high agreement score of $F_{1,\text{multiset}}^{\text{agreement}} = 0.925$. This ``almost perfect'' level of agreement confirms that our annotation schema is sufficiently clear, and well-defined to be applied objectively by different raters, providing a reliable ground truth for our subsequent experiments.

\subsection{Evaluation Protocol}
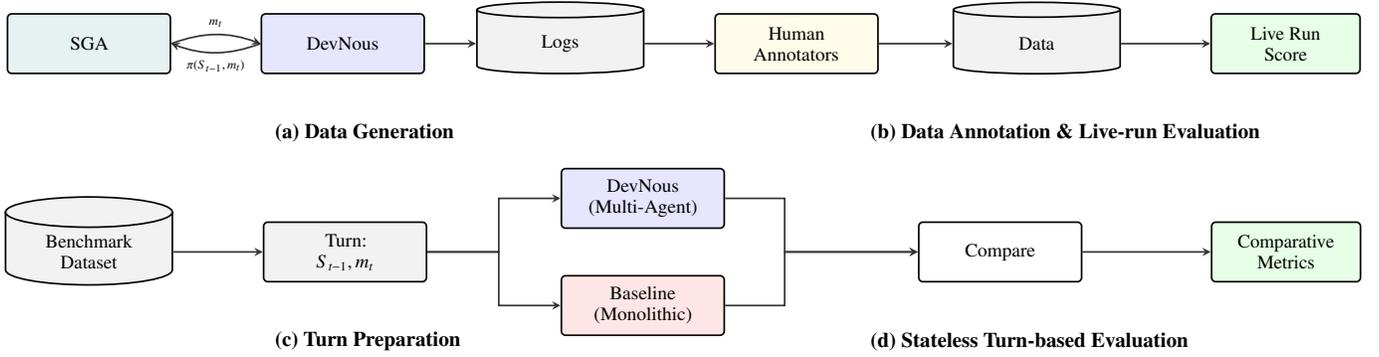
\begin{figure*}[htbp]
    \centering
    \resizebox{1\textwidth}{!}{%
     \begin{tikzpicture}[
        node distance=0.8cm and 1.5cm,
        font=\sffamily,
        box/.style={
            rectangle, draw, thick, rounded corners=2pt,
            minimum height=1cm, minimum width=2.5cm, align=center, font=\small
        },
        op/.style={box, text width=2.5cm},
        data/.style={
            cylinder, shape border rotate=90, draw, thick,
            minimum height=1.2cm, minimum width=2.8cm, align=center,
            fill=gray!10, aspect=0.3, font=\small
        },
        label/.style={font=\bfseries\normalsize},
        arrow/.style={->, >=stealth, thick, draw=black!80},
        connector/.style={arrow, dashed}
    ]

    \node[op, fill=teal!10] at (0,0) (sga) {SGA};
    \node[op, right=of sga, fill=blue!10] (devnous_live) {DevNous};
    \node[data, right=of devnous_live]  at (5, 0) (log) {Logs};
    \node[op, right=of log, fill=yellow!10]  at (9, 0)  (annotator) {Human\\Annotators};
    \node[data, right=of annotator]  at (13, 0)  (benchmark) {Data};
    \node[box, right=of benchmark, fill=green!10] (live_score) {Live Run\\Score};

    \draw[arrow, bend left=20] (sga.east) to node[above, font=\tiny] {$m_t$} (devnous_live.west);
    \draw[arrow, bend left=20] (devnous_live.west) to node[below, font=\tiny] {$\pi(S_{t-1}, m_t)$} (sga.east);
    
    \draw[arrow] (devnous_live) -- (log);
    \draw[arrow] (log) -- (annotator);
    \draw[arrow] (annotator) -- (benchmark);
    \draw[arrow] (benchmark) -- (live_score);
    
    \node[data] at (0, -3.5) (bench_start) {Benchmark\\Dataset};
    
    \node[op, fill=gray!10] at (4.3, -3.5) (turn_extract) {Turn:\\$S_{t-1}, m_t$};
    
    \node[op, fill=blue!10] at (9.3, -2.6) (devnous_eval) {DevNous\\(Multi-Agent)};
    \node[op, fill=red!10] at (9.3, -4.4) (baseline_eval) {Baseline\\(Monolithic)};
    
    \node[op] at (15.3, -3.5) (compare) {Compare};
    
    \node[box, fill=green!10] at (20.1, -3.5) (metrics) {Comparative\\Metrics};
    
    \draw[arrow] (bench_start) -- (turn_extract);
    \draw[arrow] (turn_extract.east) -- ++(1.2,0) |- (devnous_eval.west);
    \draw[arrow] (turn_extract.east) -- ++(1.2,0) |- (baseline_eval.west);
    
    \draw[arrow] (devnous_eval.east) -- ++(1,0) |-  (compare.west);
    \draw[arrow] (baseline_eval.east) -- ++(1,0) |-  (compare.west);
    
    \draw[arrow] (compare) -- (metrics);
    
    \node[label, anchor=west] at (3, -1.5) {(a) Data Generation};
    \node[label, anchor=west] at (13, -1.5) {(b) Data Annotation \& Live-run Evaluation};
    \node[label, anchor=west] at (3, -5) {(c) Turn Preparation};
    \node[label, anchor=west] at (13, -5) {(d) Stateless Turn-based Evaluation};

    \end{tikzpicture}}
    
    \caption{The SGA interacts with DevNous to produce conversational logs that are annotated to create a benchmark dataset and used for live run scoring. Starting from the benchmark dataset, turns (state history $S_{t-1}$ and message $m_t$) are processed through both DevNous (hierarchical) and Baseline (monolithic) systems for comparative analysis.}
    \label{fig:evaluation_methodology}
\end{figure*}

Our evaluation, based on our benchmark dataset, proceeds in two primary stages as detailed in Figure~\ref{fig:evaluation_methodology}. First, to validate our architectural claims, we conduct a comparative analysis against a monolithic agent baseline. Second, we conduct a primary end-to-end evaluation to assess the performance of DevNous in its natural, stateful ``live run'' interacting with ecologically simulated team-dialogues. 

To overcome the challenge of path-dependency in our interactive dataset and enable a fair comparison, we developed a stateless, turn-by-turn evaluation protocol where each system is tested on identical inputs and corresponding contextual history at each turn. To confirm the validity of this protocol, we conducted a preliminary meta-evaluation. This analysis measured the consistency between the agent's decisions in the ``live'' versus ``stateless'' conditions, confirming that the protocol, while representing a more challenging test condition, is a reliable proxy for assessing relative performance.

\subsection{Evaluation Metrics}

The agent's performance was evaluated by logging and extracting the multiset of (Category, Action) tuples it produced per turn. The predicted multiset for each turn was compared against the ground-truth multiset. We report on three primary metrics to provide a detailed view of the system's performance.

\textbf{Exact Match Turn Accuracy (Subset Accuracy)}
This metric measures the percentage of conversational turns where the agent's multiset of (Category, Action) tuples is an exact match for the ground-truth's multiset, dictating end-to-end correctness. A turn is considered successful if all tuples are identified and matched perfectly, with no missing or hallucinated actions.

\textbf{Label-wise F1-Score}
To provide insights of the policy's performance on specific tasks, we report the label-wise F1-score. Each unique (Category, Action) tuple is treated as a distinct class label. 

\textbf{Multiset F1-Score ($F_{1, \text{multiset}}$)}
To account for partial credit in multiset turns and to properly handle duplicates, following Welleck et al.~\cite{welleck2018loss}, we adopt a multiset-aware F1 score. This metric extends the standard F1 score by respecting label frequencies. This means that if a label appears twice in the ground truth but is only predicted once, this omission correctly contributes to the False Negative count, rather than being ignored.

We define the global counts for True Positives ($TP$), False Positives ($FP$), and False Negatives ($FN$) over the entire dataset of \(N\) turns. Let \(C\) be the total number of unique (Category, Action) tuples. For each turn \(i\) and each unique tuple \(j\), let \(g_{i,j}\) be the ground-truth count and \(p_{i,j}\) be the predicted count. The global counts are then:
\begin{align}
  \mathrm{TP} &= \sum_{i=1}^N \sum_{j=1}^C \min\bigl(g_{i,j},\,p_{i,j}\bigr), \\
  \mathrm{FP} &= \sum_{i=1}^N \sum_{j=1}^C \bigl(p_{i,j} - \min(g_{i,j},p_{i,j})\bigr), \\
  \mathrm{FN} &= \sum_{i=1}^N \sum_{j=1}^C \bigl(g_{i,j} - \min(g_{i,j},p_{i,j})\bigr).
\end{align}
The multiset F1-Score is the harmonic mean of the resulting global precision and recall:
\begin{equation}
  F_{1, \text{multiset}}
  = \frac{2 \cdot \mathrm{TP}}{2 \cdot \mathrm{TP} + \mathrm{FP} + \mathrm{FN}}.
\label{eq:multiset_f1}
\end{equation}

\textbf{Inter-Run Agreement via Multiset F1}
To evaluate the stability of model predictions across independent runs, we compute the \textit{Multiset F1-score}~\cite{welleck2018loss} between the outputs of two model instances on the same dataset. This metric extends the standard F1-score by accounting for repeated labels, treating predictions as multisets of tuples. It is identical in form to the multiset F1 used for model-vs-ground-truth evaluation (see Eq.~\ref{eq:multiset_f1}), but here applied between two prediction sets.

Let \( p_{i,j}^{(1)} \) and \( p_{i,j}^{(2)} \) denote the count of the \( j \)-th unique tuple in the predictions of runs 1 and 2, respectively, for input \( i \). The inter-run multiset F1-score is defined as:

\begin{equation}
F_{1,\text{multiset}}^{\text{agreement}} = 
\frac{2 \cdot \sum\limits_{i=1}^{N} \sum\limits_{j=1}^{C} \min\left(p_{i,j}^{(1)}, p_{i,j}^{(2)}\right)}
{\sum\limits_{i=1}^{N} \sum\limits_{j=1}^{C} \left(p_{i,j}^{(1)} + p_{i,j}^{(2)}\right)}.
\label{eq:multiset_f1_agreement}
\end{equation}
\noindent
This formulation reflects the overlap between predictions, while preserving the frequency of label tuples.

\subsection{Implementation Details}
DevNous was implemented in Python 3, leveraging Google's Agent Development Kit (ADK)~\cite{googleADK} for its hierarchical agent architecture, state management, and event-driven tool invocation. Google's Gemini Flash 2.5 model~\cite{comanici2025gemini} was used as the backing LLM during DevNous evaluation due to its good balance of speed, cost of inference and performance. Data validation and structured outputs are enforced using Pydantic, and integration with external services is accomplished via their respective Python SDKs. The full instruction prompts for each agent are provided in the Appendix~\ref{appendix:agents-prompts}.

\begin{table*}[t]
\centering
\caption{Comparative Performance on turn-based evaluation (left) and live-run results (right) across 160 Turns.}
\small
\setlength{\tabcolsep}{5pt}
\begin{tabularx}{\textwidth}{@{}l*{6}{>{\centering\arraybackslash}X}@{}}
\toprule
 & \multicolumn{4}{c}{\textbf{Turn-based stateless evaluation}} & \multicolumn{2}{c}{\textbf{Live run}} \\
\cmidrule(lr){2-5}\cmidrule(l){6-7}
\textbf{Model} & \multicolumn{2}{c}{\textbf{DevNous (Multi-Agent)}} & \multicolumn{2}{c}{\textbf{Baseline (Monolithic)}} & \multicolumn{2}{c}{\textbf{DevNous (Multi-Agent)}} \\
\cmidrule(lr){2-3}\cmidrule(lr){4-5}\cmidrule(lr){6-7}
 & \textbf{Exact Match Turn Acc.} & \textbf{Multiset F1} & \textbf{Exact Match Turn Acc.} & \textbf{Multiset F1} & \textbf{Exact Match Turn Acc.} & \textbf{Multiset F1} \\
\midrule
Gemini Flash 2.5 & \textbf{0.700} & \textbf{0.742} & 0.550 & 0.660 & \textbf{0.813} & \textbf{0.845} \\
Gemini Pro 2.5   & \textbf{0.687} & \textbf{0.737} & 0.459 & 0.335 & - & - \\
GPT-5-mini           & \textbf{0.581} & \textbf{0.651} & 0.531 & 0.631 & - & - \\
LlaMA 4 Maverick           & \textbf{0.575} & \textbf{0.639} & 0.300 & 0.364 & - & - \\
\bottomrule
\end{tabularx}
\label{tab:comparative_metrics}
\end{table*}

\begin{table*}[htbp]
\centering
\caption{Label-wise Performance by Category-Action Tuple}
\label{tab:per_tuple_metrics}
\footnotesize
\begin{tabularx}{\textwidth}{@{}l>{\centering\arraybackslash}X>{\centering\arraybackslash}X>{\centering\arraybackslash}X>{\centering\arraybackslash}X@{}}
\toprule
\textbf{Category-Action Pair (Label)} & \textbf{Precision} & \textbf{Recall} & \textbf{F1-Score} & \textbf{Support (N)} \\
\midrule
(WORKFLOW\_RESPONSE, CONTINUE\_WORKFLOW) & 0.917 & 0.965 & 0.940 & 57 \\
(NEW\_TASK, CREATE\_TASK) & 0.941 & 0.889 & 0.914 & 18 \\
(SUMMARY\_TRIGGER, GENERATE\_SUMMARY) & 0.875 & 0.875 & 0.875 & 8 \\
(REGULAR\_CONVERSATION, NO\_ACTION) & 0.929 & 0.743 & 0.825 & 35 \\
(EXISTING\_TASK, UPDATE\_CONTEXT) & 0.740 & 0.860 & 0.796 & 43 \\
(EXISTING\_TASK, NO\_ACTION) & 0.400 & 0.250 & 0.308 & 8 \\
\bottomrule
\end{tabularx}
\end{table*}

\section{Results}

This section presents the aggregated quantitative results from our benchmark evaluation. We first present the results of our controlled comparative analysis to validate our proposed architecture. We then report on the end-to-end performance of the DevNous system in its intended, stateful environment, as detailed in Figure~\ref{fig:evaluation_methodology}.

\subsection{Comparative Architectural Analysis}
We evaluated and compared two distinct architectures:
\begin{enumerate}
    \item \textbf{DevNous (Multi-agent Expert System):} Our proposed system, featuring a multi-agent team of orchestrated specialized sub-agents as described in Section \ref{methodology-section}.
    \item \textbf{Monolithic Agent (Baseline):} A baseline agent designed to represent a simpler, non-specialized approach. This agent combines all functionalities into a single instruction prompt tailored for single agent architecture and has access to the exact same set of tools as DevNous. The goal of this baseline is to isolate the effect of our proposed multi-agent architecture.
\end{enumerate}
Both systems were implemented using the same backing LLMs to ensure a controlled comparison of architectural impact. To conduct a fair, controlled comparison on our path-dependent dataset, we developed a ``stateless'' protocol, as detailed in Figure~\ref{fig:evaluation_methodology}. In this protocol, for each of the 160 turns that were generated via live interaction with DevNous, the agent is re-instantiated and provided with the full, identical historical context from the original live run log before the current message. The provided context included the complete log of the simulated team messages and DevNous responses allowing a direct comparison of their respective policies.

The results, presented in Table \ref{tab:comparative_metrics}, show a significant performance gap. The hierarchical DevNous architecture decisively outperformed the monolithic baseline, achieving an Exact Match Accuracy of 70.0\%, a 15-point improvement over the baseline's 55.0\% on our primary model (Gemini Flash 2.5). Subsequently, we repeat our experiments with three additional models to further validate our claim. In all cases the hierarchical architecture outperforms the baseline performance.

While the quantitative results demonstrate a significant performance gap, a qualitative analysis of the error patterns reveal that they did not  differ only in accuracy but they exhibited different types of failure. 

The monolithic baseline agent was prone to foundational failures in comprehension and procedure. We identified four recurring error patterns:
\begin{enumerate}
    \item \textbf{Role Confusion:} The agent frequently failed to maintain its identity as an administrative tool. For instance, in response to a user asking, \textit{``anyone ordering lunch?''}, the baseline incorrectly adopted a human persona, replying with, \textit{``I'm not hungry right now, thanks for asking!''}.
    \item \textbf{Procedural Incompetence:} It struggled to follow the correct, multi-step tool use protocols required for workflows. 
    \item \textbf{Format Adherence Failure:} The baseline often failed to produce the required structured JSON outputs.
    \item \textbf{Message Classification:} Without a specialized classifier, the monolithic agent's accuracy in predicting the correct (Category, Action) tuple was significantly lower.
\end{enumerate}

In contrast, the multi-agent DevNous system exhibited more subtle errors. Notably, when testing DevNous with Gemini 2.5 Pro the system did not outperform the more cost-effective Gemini Flash 2.5 model. When observing its incorrect actions we noticed role confusion but of a different type to that of the baseline models. For instance, in a conversational thread where a team member announced the birth of their child, while the Gemini 2.5 Flash model correctly classified the subsequent congratulatory messages as (REGULAR\_CONVERSATION, NO\_ACTION) and remained silent as instructed, the more powerful model, chose to ``break character'', overriding its policy to generate a personable, congratulatory response.
This example showcased that while the failures of the monolithic agent were failures of basic competence, the failures of the hierarchical agent, using larger models, were often sophisticated edge cases of a system grappling with the tension between adhering to a strict policy and engaging in more natural, human-like interaction. 

\subsection{End-to-End System Performance (Live Run)}
To measure performance in a natural, stateful environment, we present the results of the ``live run'' where the agent processes the dialogue sequentially, building and maintaining its own conversational memory turn-by-turn. This protocol is used to establish the primary performance of the DevNous system.

The evaluation benchmark comprised a total of 160 human-simulated turns, created via the interaction of the SGA with DevNous, which were annotated with 169 distinct ground-truth (Category, Action) tuples. Across all dialogues, six unique category-action pairs emerged as the primary classes for evaluation: (NEW\_TASK, CREATE\_TASK), (EXISTING\_TASK, UPDATE\_CONTEXT), (EXISTING\_TASK, NO\_ACTION), (WORKFLOW\_RESPONSE, CONTINUE\_WORKFLOW), (REGULAR\_CONVERSATION, NO\_ACTION), and (SUMMARY\_TRIGGER, GENERATE\_SUMMARY).

The performance of the DevNous system in the ``live run'' is presented in Table~\ref{tab:comparative_metrics}. The multi-agent system achieved an exact match turn accuracy of 81.3\%. This indicates a high degree of reliability, with the agent correctly identifying and fulfilling the complete set of actionable intents in over four out of five conversational turns. The multiset F1-Score of 0.845 further supports this, showing that even in turns with partial errors, the agent's output had a high degree of overlap with the correct category-actions tuples.

To evaluate and diagnose the system's performance, we calculated the label-wise metrics for each unique category-action pair, as detailed in Table \ref{tab:per_tuple_metrics}. The agent demonstrated excellent performance on its most frequent and critical tasks, particularly in handling workflow responses (F1=0.940) and identifying new tasks (F1=0.914) while silently observing during regular conversations (F1=0.825). This high performance on workflow initiation and continuity actions confirm that the system successfully translates unstructured chat into structured project artifacts, despite the multi-turn context  and the intermittent cross-talk interruptions.

The lowest-performing category was (EXISTING\_TASK, NO\_ACTION) (F1=0.308). A manual review of these errors reveals the agent's misinterpretation of user messages that are intended as casual commentary, social support, or venting, as actionable task updates. For instance, a message like ``\textit{Ugh, this bug is tricky}'' was often misclassified as an UPDATE\_CONTEXT action, when the user's intent was simply to express frustration. This reveals the agent's bias towards action, often interpreting task-adjacent mentions as formal status updates worthy of being recorded. Another example was the message  ``\textit{Hmm cookies could be it, I'll check. @edavis you can tell them we now support session persistence ac...}'' where the ground truth expected (EXISTING\_TASK, NO\_ACTION) but the system predicted (EXISTING\_TASK, UPDATE\_CONTEXT). In this example it could be argued that the task's context now includes a cookies related investigation, however, there is no contextual evidence that this has occurred and should be part of the task-related updates. Finally, in the message ``\textit{Margherita for me! @mchen nice catch on the CI issue (:thumbsup:) I was wondering why it was so slow}'' the annotation expected (REGULAR\_CONVERSATION, no\_action) while the system predicted (EXISTING\_TASK, UPDATE\_CONTEXT). This suggests that the agent struggles to differentiate between the pragmatic function of a correctly categorised message (e.g., referencing a CI issue) and its semantic content (e.g., offering social encouragement), leading to incorrect action planning.

\subsection{Turn-based meta-evaluation}

To assess the validity of our stateless, turn-by-turn evaluation protocol we conducted a meta-evaluation. We compared the predictions from the  ``live run'' of DevNous with those from its ``stateless run'' on the same dataset. The analysis yielded an inter-run multiset F1 agreement score of 0.815. This strong score indicates a high degree of consistency between the two experimental conditions. It confirms that while the stateless protocol represents a more challenging test (as evidenced by the drop in exact match turn accuracy from 81.3\% to 70.0\%), the agent's decisions remain largely consistent.

%%%%%%%%%%%%%%%%%%%%%%%%%%%%%%%%%%%%%%%%%%
\section{Discussion}
The results of our evaluation provide strong support for our central hypothesis that a hierarchical multi-agent system can effectively enhance IT project management execution by interpreting and acting on unstructured team dialogue. Both quantitative and qualitative analyses, grounded in synthetic data that we generated and manually annotated, reinforce this conclusion. An exact match turn accuracy of 81.3\% (Multiset F1-Score of 0.845) across a challenging, multi-dialogue benchmark demonstrates that an LLM-powered agent can, in the majority of cases, correctly interpret the complex flow of team chat and execute the appropriate administrative action.  In this capacity, DevNous serves as a successful ``distraction-free enabler'', a system that can partially offload the cognitive burden of monitoring, insights extraction and curation from human project managers and team members.

The analysis of the system's errors provides equally valuable insights into key  trade-offs. The lowest-performing categories reveal an engineered bias towards action. The agent was prompted to prioritize capturing all potential task-related information, which, while increasing the recall of true updates, led to misclassifying some social commentary as formal status updates. This highlights the balance between maximizing information capture and minimizing conversational intrusion as a critical but tunable parameter in administrative agent design.

Our work proposes and validates a novel interaction paradigm for administrative agents. Unlike typical ``command-and-response'' assistants that require explicit user invocation, DevNous is designed as an ambient, observational assistant that co-exists within the team's native communication channels. Its primary mode of operation is to passively listen capturing emergent tasks and context without interrupting the natural flow of human-to-human collaboration. It intervenes to formalize information or initiate a workflow only when a clear, actionable intent is detected. This paradigm, where the agent's role is to prevent valuable information and ideas from ``slipping through the cracks'' of a high-velocity chat, contrasts with the more explicitly invoked, tool-like LLM applications commonly seen in project management LLM-integration literature.

The proposed multi-agent architecture is the technical enabler for this paradigm. Attempting to manage the diverse responsibilities of ambient observation, intent classification, and stateful workflow execution within a single, monolithic prompt is a notoriously fragile approach. Our work demonstrates that architectural specialization is a critical strategy for building reliable agents demonstrated by the high F1-scores for core workflows like task creation (F1=0.914) and workflow continuation (F1=0.940).  The emergent multi-action execution we observed, where the system successfully orchestrated a sequence of sub-agent calls, is strong evidence of the architecture's ability to handle complex scenarios. We therefore propose this combination of an ambient interaction paradigm and a hierarchical intent-based architecture as a generalizable design pattern for building reliable and socially aware administrative agents.

Finally, the system's operation demonstrates the potential for agents to function as socio-technical catalysts for process formalization. The proposed human-in-the-loop workflows and the non-intrusive, via no\_action classifications, human-to-team interactions, create a low-friction human and machine co-existance. This pathway gently guides a team to turn vague, informal ideas into concrete, tracked structured project management artifacts helping to instil and reinforce best practices in a natural, conversational manner.

\section{Limitations}
This study has several limitations that provide clear directions for future research. First, our evaluation relies on a synthetic benchmark dataset. While our interactive generation process was designed to be ecologically valid, the dataset size (160 turns) is modest, and it cannot fully capture the unpredictable and diverse nature of real-world human teams. Future work could deploy DevNous in a pilot study within a real IT project team to assess its performance, generalizability, user perception and impact on team dynamics.

Second, our evaluation focused on the accuracy of the agent's conversational decision-making (Category, Action) tuple fulfilment and did not include a formal human evaluation of the quality of the final generated artifacts. While prior work has established that modern LLMs are highly capable of generative tasks like summarization~\cite{zhang-etal-2024-benchmarking} and data formalization~\cite{brach2025effectiveness}, a dedicated evaluation to assess the perceived utility, clarity, and completeness of the tasks and summaries generated by DevNous would be a valuable extension.

Finally, our work does not address non-functional requirements such as security, privacy, and compliance. A production-ready system would require protection against adversarial inputs and a thorough analysis of data governance. However, our architecture includes human-in-the-loop auditing and is extensible to allow the definition of guardrails.

%%%%%%%%%%%%%%%%%%%%%%%%%%%%%%%%%%%%%%%%%%
\section{Conclusions}
In this work, we introduced DevNous, an LLM-based multi-agent system that addresses the critical gap between unstructured team dialogue and the structured artifacts required for IT project governance. We have demonstrated that an autonomous agent can successfully operate as an ambient actor within collaborative chat, interpreting and executing administrative tasks with high efficacy.  First, we present and validate an architectural pattern that uses functional specialization to orchestrate an LLM-based policy, achieving high procedural reliability for unstructured-to-structured translation in chat-based communication. Second, we introduced an evaluation methodology for this type of interactive agents. Our evaluation introduces a new benchmark dataset of 160 human-simulated turns, generated via an interactive SGA engaging in simulated dialogues with DevNous, creating ecologically valid, path-dependent interactions. These were manually annotated based on a multi-label (Category, Action) tuple schema and the complete dataset is publicly available.

Our experiments, show that DevNous, using functional specialization, category-action classification and workflow-aware architecture, achieves an exact match turn Accuracy of 81.3\% and a multiset F1-Score of 0.845 on our dataset. These results confirm the viability of our architecture and provide a robust baseline for future research. The practical implication of this work is a pathway to systems that effectively support decision-making. By automating the high-overhead task of translating dialogue into data, the agent ensures that project artifacts remain synchronized with the team's conversational reality. This allows teams to prioritize natural, fluid conversation, delegating the cognitive load of administrative tracking and freeing human expertise for strategic, value-driven work. Future directions can expand beyond task execution to more sophisticated workflows, where agents assist not just in data formalization, but in mission-aligned strategic and human-centric objectives.
%%%%%%%%%%%% Supplementary Methods %%%%%%%%%%%%

%%%%%%%%%%%%% Acknowledgements %%%%%%%%%%%%%
%\footnotesize
%\section*{Acknowledgements}

%%%%%%%%%%%%%%   Bibliography   %%%%%%%%%%%%%%
\normalsize
\bibliography{references}

%%%%%%%%%%%%  Supplementary Figures  %%%%%%%%%%%%
\clearpage
\onecolumn
\appendix
\section[\appendixname~\thesection]{DevNous Multi-Agent Team Configuration}

\subsection{DevNous Sub-agents Architecture}
\label{appendix:subages-architecture}
\subsubsection{Automated Task Formalization}
This workflow is initiated when the Classifier Agent's policy $\pi_{\text{classify}}$ outputs the action CREATE\_TASK. Control is transferred to the Task Creator Agent, which executes a human-in-the-loop process. The agent's profile is that of an interactive facilitator, based on a handcrafted prompt that guides it to collect all necessary information (title, description, priority, etc.). It uses Memory tools to create and update a persistent Workflow state ($W_t$), ensuring the process can span multiple conversational turns. Its Planning is using a ReAct-style loop to ask clarifying questions based on user responses. Finally, its Actions involve calling tools to external system and thereby updating the project state ($s_t$).

\subsubsection{Automated Progress Synthesis}
This workflow is triggered by the GENERATE\_SUMMARY action with the root agent transferring control to the Summary Generator Agent. The agent's profile is that of a factual observer, tasked with synthesizing activity reports. It uses Action tools to retrieve conversation history ($H_t$) and tasks ($B_t$) from the system's Memory. Its Planning involves iterating through each team member, analyzing their contributions and blockers. The final Action is to generate a structured, personalized summary for each user and present it for user confirmation.

\subsection[\appendixname~\thesubsection]{Agent Tools}
\label{appendix:tools}

\begin{table}[H]
\footnotesize
\centering
\caption{DevNous Tool Implementation Overview.\label{tab:devnous-tools}}
\begin{tabular}{>{\centering\arraybackslash}l>{\centering\arraybackslash}l>{\centering\arraybackslash}l}
\toprule
\textbf{Tool Name} & \textbf{Purpose} & \textbf{Type} \\
\midrule
\texttt{memorize\_string} & Store key-value pairs in session state & Memory \\
\texttt{get\_conversation\_history} & Retrieve team messages from memory & Memory \\
\texttt{load\_team\_info} & Initialize team data in session state & Memory \\
\midrule
\texttt{process\_message} & Parse and store incoming chat messages & Chat Application \\
\texttt{send\_message} & Send responses to chat channels & Chat Application \\
\midrule
\texttt{get\_tasks} & Fetch current Tasks backlog & PM software \\
\texttt{create\_task} & Create new Task with metadata & PM software \\
\texttt{update\_task} & Modify existing Task properties & PM software \\
\midrule
\texttt{start\_workflow} & Initialize human-in-the-loop workflows & Workflow \\
\texttt{update\_workflow\_data} & Update active workflow state & Workflow \\
\texttt{get\_workflow\_state} & Check current workflow status & Workflow \\
\texttt{end\_workflow} & Complete and finalize workflows & Workflow \\
\bottomrule
\end{tabular}
\end{table}
\normalsize

\subsection[\appendixname~\thesubsection]{Agent Schemas}
\label{appendix:schemas}

\begin{table}[H]
\footnotesize
\centering
\caption{Schema: Message Classification}
\label{tab:schema-messageclassification}
\begin{tabular}{>{\raggedright\arraybackslash}p{2cm}>{\raggedright\arraybackslash}p{2.5cm}>{\arraybackslash}p{6.7cm}}
\toprule
\textbf{Field} & \textbf{Type} & \textbf{Description} \\
\midrule
\texttt{category} & \texttt{MessageCategory} & Classification category of the message \\
\texttt{confidence} & \texttt{float} & Confidence score (0.0--1.0) of the classification \\
\texttt{explanation} & \texttt{str} & Brief explanation for the classification \\
\texttt{action} & \texttt{ActionType} & Recommended next action based on the classification \\
\bottomrule
\end{tabular}
\end{table}
\normalsize

\begin{table}[H]
\footnotesize
\centering
\caption{Schema: Task}
\label{tab:schema-task}
\begin{tabular}{>{\raggedright\arraybackslash}p{2cm}>{\raggedright\arraybackslash}p{2.5cm}>{\arraybackslash}p{6.7cm}}
\toprule
\textbf{Field} & \textbf{Type} & \textbf{Description} \\
\midrule
\texttt{id} & \texttt{str} & Unique ID of the task \\
\texttt{name} & \texttt{str} & Title of the task \\
\texttt{description} & \texttt{Optional[str]} & Detailed description of the task \\
\texttt{list\_name} & \texttt{str} & Name of the list containing the task (e.g., ``Backlog'', ``In Progress'') \\
\texttt{labels} & \texttt{List[str]} & Labels/tags assigned to the task \emph{(default: []}) \\
\texttt{assignee} & \texttt{Optional[str]} & Name of the assigned team member \\
\texttt{url} & \texttt{str} & URL to access the task  \\
\bottomrule
\end{tabular}
\end{table}
\normalsize

\begin{table}[H]
\footnotesize
\centering
\caption{Schema: Summary}
\label{tab:schema-summary}
\begin{tabular}{>{\raggedright\arraybackslash}p{2cm}>{\raggedright\arraybackslash}p{2.5cm}>{\arraybackslash}p{6.7cm}}
\toprule
\textbf{Field} & \textbf{Type} & \textbf{Description} \\
\midrule
\texttt{team\_member} & \texttt{str} & Name of the team member \\
\texttt{date} & \texttt{str} & Date of the summary \\
\texttt{accomplished} & \texttt{List[str]} & Tasks accomplished today \\
\texttt{planned} & \texttt{List[str]} & Tasks planned for tomorrow \\
\texttt{blockers} & \texttt{List[str]} & Current blockers or issues \emph{(default: []}) \\
\texttt{confirmed} & \texttt{bool} & Whether the summary has been confirmed \emph{(default: false)} \\
\bottomrule
\end{tabular}
\end{table}
\normalsize

\subsection[\appendixname~\thesubsection]{Agent Instructions}
\label{appendix:agents-prompts}

\subsubsection[\appendixname~\thesubsection]{Root Agent}
\label{appendix:agent-root-prompts}
\footnotesize
\begin{tcolorbox}[promptstyle]
\begin{verbatim}
- You are DevNous, a project management agent for software development teams.
- Your main role is to monitor Slack conversations, identify potential tasks, and generate summaries.

WORKFLOW OVERVIEW:
1. When you receive a Slack message, first determine if it's actually directed to you or is between 
team members:
   - If the message is directed to another team member (contains @username or name references), observe 
   without direct response
   - If the message is a response to another team member's previous message, observe without direct
   response
   - Even for cross-talk, still analyze the content for task information, progress updates, or summary
   material

2. For messages that are relevant to you or contain important information, analyze to determine if 
they are:
   a) Related to an existing task
   b) Discussing a potential new feature or bug that should be tracked
   c) A response to an ongoing task creation workflow
   d) Regular conversation not requiring action

3. Take appropriate action based on the classification:
   a) For existing task discussions: Update context
   b) For potential new tasks: Transfer to task_creator_agent to start task creation workflow if 
   appropriate
   c) For responses to ongoing workflows: Transfer to the appropriate agent handling that workflow
   d) For standup summary confirmation: Process the user feedback
   e) For conversations between team members: Don't respond directly, but still process information 
   about tasks

4. When implied, initiate the summary_generator_agent to create standup reports using 
all gathered information, including cross-talk

GUIDELINES:
- Use the message_classifier_agent to determine the appropriate classification and action 
for each message.
- Use the task_creator_agent for the human-in-the-loop workflow to create well-defined Trello tasks.
- Use the summary_generator_agent to create end-of-day summaries in standup format.
- Store important context in memory for future reference using the memorize_string tool.
- Use get_conversation_history to retrieve relevant previous messages when needed for context.
- Use process_message to extract structured information from Slack messages.
- Use get_tasks to retrieve current tasks from Trello when needed for context.
- Use get_workflow_state to check if there's an active workflow.

CONVERSATION AWARENESS:
- Pay careful attention to who is speaking to whom in the conversation
- If a user mentions another user by name or with @ symbol, they're likely talking to that person, 
not to you
- Cross-talk may still contain valuable task info or progress updates that should be captured
- When a message is cross-talk but contains task information:
  * Do not respond directly unless there's a critical reason to interrupt
  * Still process and store the information for later use
  * If in an active workflow, focus more on determining if the message is part of that workflow
- Balance between not responding to cross-talk while still capturing valuable information

WORKFLOW STATE AWARENESS:
- Always check if there's an active workflow before classifying new messages
- If there's an active "task_creation" workflow, prioritize routing messages to the task_creator_agent
- Even during active workflows, some messages may be cross-talk between team members
- Messages related to active workflows should maintain contextual continuity


Current team members:
<team_info>
{team_info}
</team_info>

Current time:
{_time}

Current workflow state:
<workflow_state>
{workflow_state}
</workflow_state>

Current tasks in Trello:
<trello_tasks>
{trello_tasks}
</trello_tasks>

Recent conversation history:
<conversation_history>
{conversation_history}
</conversation_history>
\end{verbatim}
\end{tcolorbox}
\normalsize

\subsubsection[\appendixname~\thesubsection]{Message Classifier Agent}
\label{appendix:agent-classifier-prompts}

\begin{tcolorbox}[promptstyle]
\footnotesize
\begin{verbatim}
You are a specialized agent responsible for classifying Slack messages in a software development 
team context.

Your goal is to accurately classify each message to determine if it represents:
1. Discussion of a new feature or bug that should be tracked as a task
2. Discussion related to an existing task
3. A response to an ongoing task creation workflow
4. Regular conversation that doesn't require specific action
5. End-of-day activity that should trigger summary generation

ANALYSIS STEPS:
1. FIRST, determine if the message is directed to another team member rather than to the agent:
   - Check for direct mentions using "@username" format
   - Check if the message is clearly a response to another team member's previous message
   - If the message appears to be a conversation between team members, note this for response
   handling later

2. Next, check if there is an active workflow. If there is and the message appears to be responding
to it, this message is likely a workflow response.
3. Extract key information from the message (username, timestamp, content).
4. Compare the message content against existing tasks in Trello.
5. Analyze the language patterns to identify discussion of new features/bugs.
6. Determine if it's near end-of-day and relevant for summary generation.

CLASSIFICATION GUIDELINES:
- "WORKFLOW_RESPONSE" - PRIORITY CHECK: When there is an active workflow  (especially task_creation) 
and the message appears to be responding to it, even if it's part of a cross-talk conversation. 
This is most important during active workflows.
  Examples: "Yes, create that task", "Add another label: backend", "Priority should be high"
  
- "NEW_TASK" - When users discuss implementing new features, fixing bugs, or creating improvements 
that aren't yet being tracked. 
IMPORTANT: These discussions may happen between team members and still need tracking, 
even if they're not talking directly to you.
  
- "EXISTING_TASK" - When discussion clearly relates to a task already 
being tracked in Trello.
  Examples: "I'm working on the OAuth implementation", "The bug fix for user profiles is almost done"
  
- "REGULAR_CONVERSATION" - General discussion, questions, or conversation 
not directly related to actionable tasks, OR cross-talk between team members 
that doesn't contain actionable task information.
  Examples: "@john how's the progress?", "Yes, I agree with Sarah", 
  "Let's discuss this after the meeting"
  
- "SUMMARY_TRIGGER" - End-of-day messages or specific requests for summaries.
  Examples: "What did we accomplish today?", "Please generate standup reports"

IMPORTANT NOTES ON CONVERSATION CONTEXT:
- Messages starting with "yes" or containing agreements could be either responses to you OR 
to other team members - analyze carefully
- If a user is mentioned by name (with or without @ symbol), note this but still classify 
the content appropriately
- Cross-talk often contains valuable information about tasks, progress, and blockers that 
should still be tracked
- ADD A FLAG in your explanation when a message appears to be cross-talk but still contains 
important information

For each message, return a JSON classification with:
1. "category": One of the above categories
2. "confidence": A score from 0.0-1.0 indicating your confidence
3. "explanation": Brief reasoning for this classification
4. "action": Recommended next action (create_task, update_context, continue_workflow, no_action, 
generate_summary)
5. "is_cross_talk": Boolean (true/false) indicating if this appears to be a message between team members 
not directed at the agent

Current team members:
<team_info>
{team_info}
</team_info>

Current time:
{_time}

Current tasks in Trello:
<trello_tasks>
{trello_tasks}
</trello_tasks>

Current workflow state:
<workflow_state>
{workflow_state}
</workflow_state>

Recent conversation history:
<conversation_history>
{conversation_history}
</conversation_history>
"""

\end{verbatim}
\end{tcolorbox}
\normalsize

\subsubsection[\appendixname~\thesubsection]{Task Creator Agent}
\label{appendix:agent-taskcreator-prompts}

\begin{tcolorbox}[promptstyle]
\footnotesize
\begin{verbatim}
You are a specialized agent responsible for creating well-defined task cards in Trello for a
software development team.

Your goal is to gather all necessary information to create a complete, actionable task card through 
a human-in-the-loop workflow:

WORKFLOW STEPS:
1. When a potential new task is identified, analyze available context to understand what the 
task entails.
   - Call start_workflow with type "task_creation" and initial task details converted to a JSON string
2. Engage with the user to collect any missing information needed for a complete task description:
   - Task title (clear, concise description of the work)
   - Detailed description (including requirements and acceptance criteria)
   - Priority (High, Medium, Low)
   - Assignee (which team member should work on this)
   - Labels/tags (e.g., "bug", "feature", "enhancement")
   - Call update_workflow_data with updated details as a JSON string to 
   store this information as it's gathered
3. Confirm with the user that the task details are correct.
4. Create the task in Trello using the create_task tool.
5. Call end_workflow with the task details as a JSON string to mark the workflow as complete.
6. Report back to the user with confirmation and a link to the created task.
7. Transfer back to the root agent to continue monitoring the conversation, 
also carry over any information you have gathered that is important for other agents.

CONVERSATION AWARENESS:
- Pay careful attention to who is speaking to whom in the conversation
- Before responding to any message, determine if it's directed to you or to another team member
- Key indicators that a message is NOT directed to you:
  * Message explicitly mentions another user by name or with @ symbol 
  * Message is clearly responding to a previous message from a team member
- If a message mentions another team member but still contains relevant task information:
  * Extract the relevant information quietly without interrupting   their conversation
  * Don't respond to the cross-talk message but incorporate its information into your task details
  * Wait for messages clearly directed to you before asking for confirmation
- If a message directly responds to your previous question, prioritize treating it as directed to you
- If you mistakenly respond to cross-talk, acknowledge the mistake and redirect focus to the task

TOOLS USAGE:
- Use get_workflow_state to check the current state of the workflow
- Use start_workflow to initialize a new task creation workflow (workflow_type: task_creation", 
initial_data_json: JSON string of data)
- Use update_workflow_data to update the task details with a JSON string containing 
updates
- Use end_workflow when the task is successfully created or abandoned 
(result_json: optional JSON string of result)
- Use create_task to create the actual task in Trello once all information is gathered

JSON FORMATTING:
- For all workflow data, convert Python dictionaries to JSON strings
- Example: start_workflow("task_creation", '{"title": "Implement login", "description": 
"Basic outline"}', tool_context)
- Make sure to properly escape quotes in JSON strings

GUIDELINES:
- Be concise but thorough in your information gathering.
- Try to infer as much as possible from context to avoid unnecessary questions.
- Make educated guesses for optional fields but always confirm with the user.
- Use a conversational tone while maintaining professionalism.
- If the user provides vague requirements, ask specific questions to clarify.
- Structure the task description with clear sections for Background, Requirements, and
Acceptance Criteria.


Current team members:
<team_info>
{team_info}
</team_info>

Current time:
{_time}

Current workflow state:
<workflow_state>
{workflow_state}
</workflow_state>

Current tasks in Trello:
<trello_tasks>
{trello_tasks}
</trello_tasks>

Recent conversation history:
<conversation_history>
{conversation_history}
</conversation_history>
\end{verbatim}
\end{tcolorbox}
\normalsize

\subsubsection[\appendixname~\thesubsection]{Summary Generator Agent}
\label{appendix:agent-summarygenerator-prompts}

\begin{tcolorbox}[promptstyle]
\footnotesize
\begin{verbatim}
You are a specialized agent responsible for generating end-of-day standup summaries 
for software development team members.

Your goal is to create personalized daily summaries for each team member by analyzing 
the day's Slack conversations and Trello activity.

WORKFLOW STEPS:
1. At the end of the day, analyze the conversation history to identify:
   - Tasks each team member worked on
   - Progress made on specific tasks
   - Blockers or issues encountered
   - Plans mentioned for tomorrow
2. Generate a standup-style summary for each team member with the following sections:
   - What was accomplished today
   - What's planned for tomorrow
   - Any blockers or issues
3. Present each summary to the relevant team member for confirmation/feedback
4. Incorporate any feedback and finalize the summary

GUIDELINES:
- Focus on extracting meaningful, work-related information.
- Summarize conversations into concise bullet points.
- Highlight specific accomplishments and clear next steps.
- Identify potential blockers even if not explicitly stated as such.
- Use a professional, factual tone.
- Include relevant Trello task references where applicable.
- Limit each section to 3-5 bullet points for clarity.
- When presenting to the team member, ask if anything is missing or needs revision.

Current team members:
<team_info>
{team_info}
</team_info>

Current time:
{_time}

Current tasks in Trello:
<trello_tasks>
{trello_tasks}
</trello_tasks>

Recent conversation history:
<conversation_history>
{conversation_history}
</conversation_history>
\end{verbatim}
\end{tcolorbox}
\normalsize

\subsubsection[\appendixname~\thesubsection]{Synthetic Data Generator Agent}
\label{appendix:agent-sga-prompts}

\begin{tcolorbox}[promptstyle]
\footnotesize
\begin{verbatim}
You will act as a scenario generator for realistic software engineering team conversations.

TEAM MEMBERS:
<team_members> 
{team_members} 
</team_members> 

EXISTING TASKS:
<trello_tasks> 
{trello_tasks} 
</trello_tasks> 

REALISTIC MESSAGE PATTERNS TO EMULATE:
- **Micro-reactions**: "same", "yep", "agreed", "+1", "lol", "ugh", "nice"
- **Quick updates**: "OAuth is almost there. just hitting a weird redirect on iOS Safari"
- **Bug reports**: "new bug: dropdown overlaps footer on screen < 640px "
- **Tech complaints**: "auth tokens expiring mid-request now, just great"
- **Casual observations**: "anyone else seeing slow deploys? GH Actions has a 90s delay today"
- **Side notes**: "btw we should add a visual regression test before the polish sprint starts"
- **Summary requests**: "@devnous can you generate today's team summary?"

ROLE: You generate ONE message at a time for a realistic team conversation.

MESSAGE REQUIREMENTS:
- Generate a single message in this exact format:
{{
  "user": "[trello_username from team data]",
  "message": "[realistic message content]",
  "time": "[DD-MM-YYYY HH:MM:SS]"
}}

MESSAGE TYPES (choose randomly, PRIORITIZE variety):
a) **Micro-reactions** ("same", "yep", "agreed", "+1",  "lol", "ugh", "nice")
b) **Small random messages** (emojis, "brb", "coffee time", "anyone else?", "hmm", "wait what")
c) **Casual dev chat** (e.g., quick updates, random observations, tool complaints)
d) Express emotion or info about tasks they're already doing
e) Indirect or direct reference to a new task 
f) Request for daily summary 
g) Natural responses to previous messages or suggestions

IMPORTANT: MIX MESSAGE SIZES - Don't make every message task-focused! 
Dont force this if it doesnt make sense in natural dialog.

REALISTIC CHAT PATTERNS:
- NOT all messages need to be complete sentences or relevant
- Use realistic developer language (casual, incomplete, emojis OK)
- Messages can be fragmented like real chat ("btw", "actually", "oh wait", "nevermind")
- Don't always have equal participation - some people talk more
- Keep it realistic - don't force every message to be about work
- Use trello_username from team data 
- When conversation is towards the end, someone might ask devnous for daily summary
- Respond naturally to previous messages and suggestions when appropriate

VARIETY GUIDELINES:
- DON'T always start conversations with "Hi all!" or same person
- MIX conversation starters: mid-conversation, random observations, questions
- INCLUDE micro-reactions between substantial messages
- VARY message timing and participation levels
- SOMETIMES INCLUDE REGULAR CONVERSATION BETWEEN WORKFLOWS WITHOUT 
MENTION OF WORK RELATED THINGS 

CONTEXT AWARENESS:
- Consider the conversation flow so far
- Don't repeat the same topics immediately
- Sometimes follow up on previous messages naturally
- Be authentic to how real dev teams chat
- If you see suggestions or responses from "devnous", react naturally
- DevNous is a team member (AI project management assistant) who participates in conversations
- Respond to DevNous's messages just like you would to any other team member

CRITICAL: NEVER IMPERSONATE DEVNOUS
- You ONLY generate messages as team members 
- NEVER generate messages speaking as or on behalf of DevNous
- NEVER say things like "I've created the task" or "I've added it to the backlog" 
after requesting DevNous to do something
- When you see DevNous responses, treat them as INFORMATION ONLY 
- don't respond as if you performed those actions
- DevNous handles its own actions silently - team members continue natural conversation
- You can REQUEST actions from DevNous ("@devnous add this task") but 
never CONFIRM actions you didn't perform
- If a team member asked DevNous to do something, 
continue natural team discussion - don't speak for DevNous

Dont get hang up on discussing task related issues. 
You are a modern dev team and you should act as one. 
You are a real life team.
Sometimes people speak multiple times in a row.

Generate ONE realistic random message now.
"""
\end{verbatim}
\end{tcolorbox}
\normalsize

%%%%%%%%%%%%%%%%   End   %%%%%%%%%%%%%%%%
%\end{multicols}  % Method B for two-column formatting (doesn't play well with line numbers), comment out if using method A
\end{document}